\documentclass[review]{elsarticle}
\usepackage{lineno,hyperref}
\usepackage{graphicx}
\usepackage{pdfpages}
\usepackage{times}
\usepackage{epsfig}
\usepackage{amsmath,amssymb,amstext}
\usepackage{amssymb}
\usepackage{float}
\usepackage{multirow}
\usepackage{algorithm}
\usepackage{algorithmic}

\usepackage{booktabs}
\usepackage{epstopdf}
\usepackage{subfigure}
\usepackage{booktabs}
\usepackage{diagbox}
\usepackage{float}
\usepackage{makecell}
\modulolinenumbers[1]
\journal{Neurocomputing}

\begin{document}
\begin{frontmatter}
	\title{Self-supervised Graph Representation Learning via Bootstrapping}
	
	\author{Feihu Che$^{1,2}$, Guohua Yang$^{1}$, Dawei Zhang$^{1}$, Jianhua Tao$^{1,2,3}$, Pengpeng Shao$^{1,2}$, Tong Liu$^{1}$}
	
	\address{$^1$National Laboratory of Pattern Recognition, Institute of Automation,  Chinese Academy of Sciences, Beijing, China \\
		$^2$School of Artificial Intelligence, University of Chinese Academy of Sciences, Beijing, China\\
		$^3$CAS Center for Excellence in Brain Science and Intelligence Technology, Beijing, China
		\\
	}

\begin{abstract}
	Graph neural networks~(GNNs) apply deep learning techniques to graph-structured data and have achieved promising performance in graph representation learning. However, existing GNNs rely heavily on enough labels or well-designed negative samples. To address these issues, we propose a new self-supervised graph representation method: deep graph bootstrapping~(DGB). DGB consists of two neural networks: online and target networks, and the input of them are different augmented views of the initial graph. The online network is trained to predict the target network while the target network is updated with a slow-moving average of the online network, which means the online and target networks can learn from each other. As a result, the proposed DGB can learn graph representation without negative examples in an unsupervised manner. In addition, we summarize three kinds of augmentation methods for graph-structured data and apply them to the DGB. Experiments on the benchmark datasets show the DGB performs better than the current state-of-the-art methods and how the augmentation methods affect the performances.
\end{abstract}

\end{frontmatter}

\section{Introduction}
Graph neural networks~(GNNs) have made remarkable advancements in representation learning for graph-structured data~\cite{kipf2016semi,gilmer2017neural,velivckovic2017graph}.
Combining modeling the rich topology of graphs and unparalled expressive ability of deep learning, GNNs learn low-dimensional embeddings from variable-size and permutation-invariant graphs. The success of GNNs has also benefited a  wide range of applications, such as in social networks~\cite{kipf2016semi}, modecules~\cite{duvenaud2015convolutional}, robot designs~\cite{wang2019neural} and knowledge graphs~\cite{vivona2019relational}.

Like other deep learning methods, many existing GNNs and their variants are mainly based on semi-supervised setting so they need a certain number of labeled data. However, requiring enough quality labeled data may meet some challenges in the real application scenarios. For instance, biology graphs represent specific concepts~\cite{sun2019infograph}, so it is difficult and expensive to annotate the graphs; in addition, the reliability of the given labels may sometimes be questionable~\cite{peng2020graph}.
%""

Though the present unsupervised graph representation algorithms do not need labels, they rely heavily on negative samples. For example, random walk-based methods~\cite{perozzi2014deepwalk,tang2015line} consider node pairs that are "close"  in the graph are positive samples, meanwhile, take node pairs that are "far" in the graph as negative samples. The loss function enforces "close" node pairs to have more similar representations than "far" node pairs. In addition, deep graph informax~(DGI)~\cite{velickovic2019deep} maximizes mutual information between patch representations and corresponding high-level summaries while taking the corruption graph as negative samples. The performances of these methods are highly dependent on the choices of negative samples, but in the wild negative pairs are not easy or computationally expensive to acquire. Consequently, how to obtain high-quality graph representation without supervision or negative examples becomes necessary for a number of practical applications, which motivates the study of this paper.

Recently, BYOL~(bootstrap your own latent)~\cite{grill2020bootstrap} introduces bootstrapping mechanism to visual representation learning and learns from its previous version, which has achieved state-of-the-art results without negative examples. Nevertheless, BYOL is based on image data, and to the best knowledge of us, no work has applied bootstrapping mechanism to graph-structure data. In this work, we extend BYOL to graph-structured data and propose deep graph bootstrapping~(DGB). DGB relies on two neural networks: online and target networks. During one time training, the target network is fixed, then the online network is updated by gradient descent to predict the target network; after training, the target network parameters are updated with a slow-moving average of the online network parameters. Such training mechanism can make the online and target networks learn from each other, hence, DGB no longer needs labeled data or negative examples.
%这一段图数据增强要不要和图像数据增强比一比，说明其难度。

Data augmentations play an important role in DGB, but how to design efficient augmentation methods for graph-structured data is still challenging. In this paper, we systematically summarize three kinds of augmentation methods: node augmentation, including node feature dropout and node dropout; adjacent matrix augmentation, including personalized PageRank~(PPR)~\cite{page1999pagerank} and heat kernel~\cite{kondor2002diffusion}; and the combination of them. We also apply these augmentation methods to DGB and show how augmentation methods help to improve the performances of DGB.

In conclusion, we propose a new unsupervised graph representation learning method without negative examples and systematically summarize three augmentation methods for graph-structured data. The main contributions of this paper are as followings:

\begin{itemize}
	\item We first generalize bootstrapping mechanism to graph-structured data, and propose an unsupervised graph representation learning method DGB without negative examples.
	\item The experimental results of the benchmark datasets show the DGB model is superior to the present supervised and unsupervised graph representation models.
	\item We systematically summarize three kinds of augmentation methods for graph-structured data, and apply them to DGB then analyze how data augmentations affect the performances of DGB.
	
\end{itemize}

%Nevertheless, it is not clear how to apply this technique to graph-structured data. To address this, in this paper, we propose a new self-supervised approach for unsupervised graph representation learning that is based on bootstrapping latent representations, without using negative pairs. The idea is adapted from BYOL~\cite{grill2020bootstrap}, which trains two neural networks(online and target network) and learn from each other.

\section{Related Works}

\subsection{Random Walk Based Methods}
Random walk-based methods~\cite{perozzi2014deepwalk,tang2015line,grover2016node2vec} generate random walks across nodes, and then apply neural language modes  to get network embedding. They are known to over-emphasize proximity information at the expense of structural information~\cite{velickovic2019deep,hassani2020contrastive}.
\subsection{Target Networks}
Target networks have a wide range of applications in deep reinforcement learning~\cite{van2018deep}. As one of the two important components of deep Q-network~\cite{mnih2015human}, target networks make the training process more stable and alleviate oscillations or divergence. In deep Q-network, every C updates the network $Q$ is cloned to obtain a target network $\hat{Q}$, and  $\hat{Q}$ is used to generate the Q-learning targets for the following C updates. Target networks are extended to soft target updates, rather than directly copying the weights in \cite{lillicrap2015continuous}, as a result, the target values have to change slowly, which can improve the stability of learning. 
\subsection{Graph Diffusion}
Based on sparsified generalized graph diffusion, graph diffusion convolution(GDC) is a powerful spatially extension of message passing in GNNs~\cite{klicpera2019diffusion}. GDC can generate a new sparse graph which is not limited to message passing neural network, so GDC is able to be applied for any existing graph-based model or algorithm without requiring  changing the model.  
\subsection{Contrastive Methods}
Contrastive methods employ a scoring function that enforces a higher score on positive pairs and a lower score on negative pairs~\cite{velickovic2019deep,li2019graph}. Some recent works adapt contrastive ideas in image representation learning to unsupervised graph learning~\cite{velickovic2019deep,sun2019infograph}. Deep graph informax~\cite{velickovic2019deep} extends deepInfomax(DIM)~\cite{hjelm2018learning} and contrasts node and graph embedding to learn node embeddings; in addition, InfoGraph extends DIM to learn graph embeddings.

\subsection{Bootstrapping Methods}
Different from contrastive methods' requiring many negative examples to work well~\cite{he2020momentum,chen2020simple}, bootstrapping methods can learn representations without negative examples in an unsupervised manner. DeepCluster~\cite{caron2018deep} produces targets for the next representation by bootstrapping the previous representation; it clusters data points based on the prior representation and uses the clustered index of each example as a classification target to train the new representation. Predictions of Bootstrapped Latents(PBL)~\cite{guo2020bootstrap} apply bootstrapping methods to multitask reinforcement learning. PBL predicts latent embeddings of future observations to train its representations, and the latent embeddings are themselves trained to be predictive of the aforementioned representations. BYOL~\cite{grill2020bootstrap} uses two neural networks, referred to as online and target networks; the outputs of the target network serve as targets to train the online network and the parameters of the target network are updated by a slow-moving average of the online network parameters.

\begin{figure}[t]
	\centering
	\includegraphics[width=1.\columnwidth]{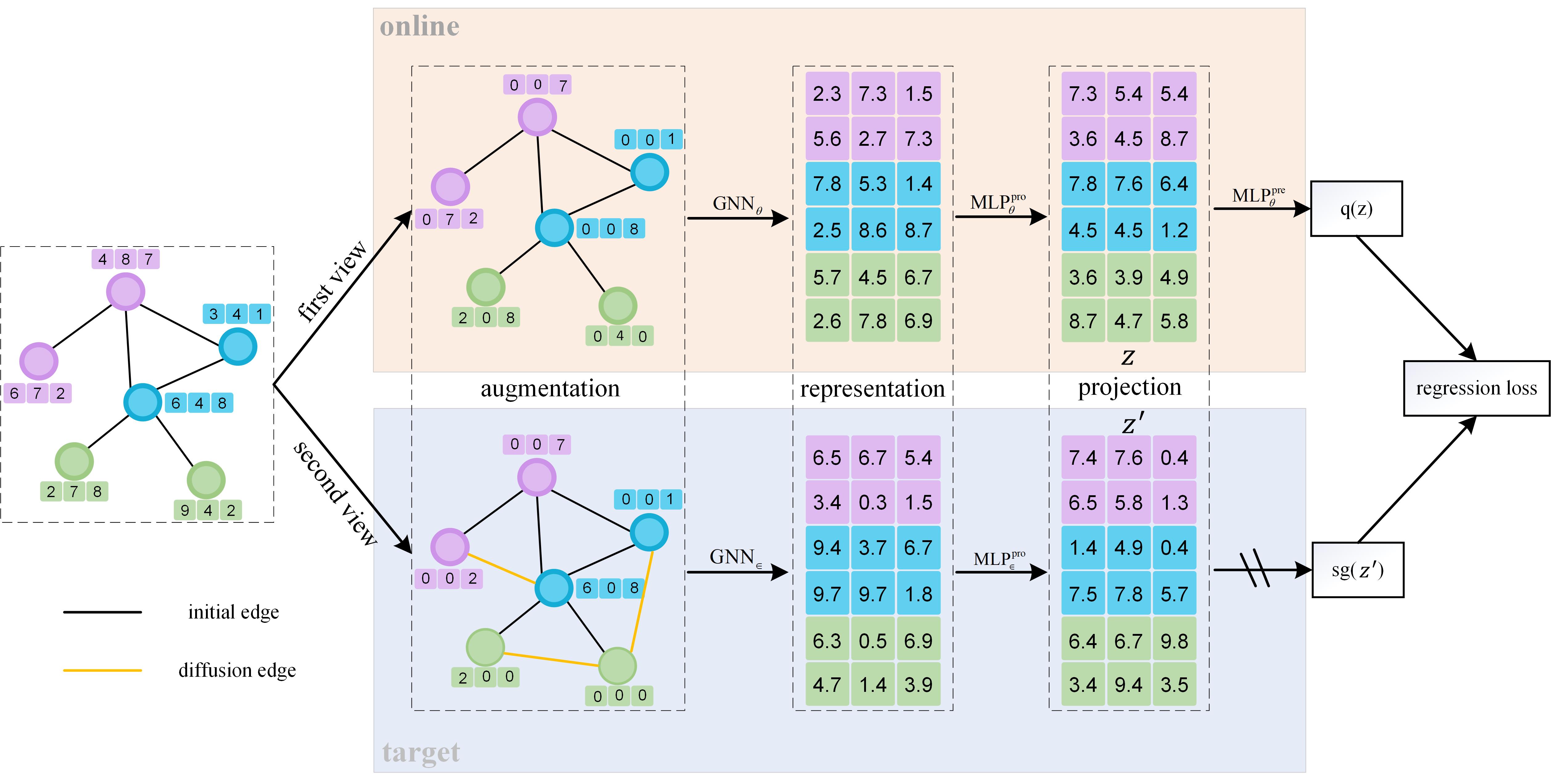} 
	\caption{The proposed deep graph bootstrapping model for graph representation learning. The model consists of two neural networks: online and target networks. The input of online and target networks are two augmented views of the initial graph. We feed the first view to a graph convolution network~(GCN) in order to obtain node representations, then the representations are fed to two MLPs successively to get the projections and predictions; for the target network, the second view is fed to a GCN and one MLP to get representations and projections. The objective function of the model is to minimize the regression loss between the predictions of the online network and the projections of the target network. To note that, the parameters of the target network are not updated by gradient descent. After training, we only use the GCN layer of the online network to get node representations for downstream tasks.}
	\label{fig_whole}
\end{figure}

\section{Unsupervised Graph Representation Learning}
A graph can be represented as $\mathcal{G} = \{\mathbf{X}, \mathbf{A}\}$, where $\mathbf{X}=\left\{\vec{x}_{1}, \vec{x}_{2}, \ldots, \vec{x}_{n}\right\}$ represents the node features, $n$ is the number of nodes in the input graph and $\vec{x}_{i} \in \mathbb{R}^{d}$ means the feature vector of node $i$; $\mathbf{A} \in \mathbb{R}^{n \times n}$ is an adjacency matrix, $A_{i j}=1$ represents there exists an edge from node $i$ to node $j$ and $A_{i j}=0$ otherwise.

The objective of DGB is to learn an encoder, $\mathcal{E}: \mathbb{R}^{n \times d} \times \mathbb{R}^{n \times n} \rightarrow \mathbb{R}^{n \times d^{\prime}}$, as a result we can get $\mathbf{H}=\mathcal{E}(\mathbf{X}, \mathbf{A})=\left\{\vec{h}_{1}, \vec{h}_{2}, \ldots, \vec{h}_{n}\right\}$, $\vec{h}_{i} \in \mathbb{R}^{d^{\prime}}$ represents the final learned embedding for node $i$, which is for the  downstream tasks, such as node classification or node cluster.

\section{Deep Graph Bootstrapping}
In this section, we introduce the proposed DGB model in detail. DGB model is inspired by BYOL~\cite{guo2020bootstrap}, which learns visual representations by bootstrapping the latent representations. The DGB model learns node embeddings by predicting previous versions of its outputs, without leveraging negative examples. 
As is shown in figure1, DGB refers to two neural networks, online network and target network. The online network is trained to predict the target network output, and the target network is updated by an exponential moving average of the online network~\cite{lillicrap2015continuous}. To be specific, DGB model consists of the following components:
\begin{itemize}
	\item Graph convolution network as graph encoders to get node representations.
	\item Data augmentations for graph-structured data including node augmentation and graph diffusion.
	\item A bootstrapping mechanism to make the online network and target network learn from each other.
\end{itemize}

\subsection{Graph Neural Network Encoder}
GNNs learn representations through an iterative process of transforming and aggregating from topological neighbors.
In this paper, for simplicity and generalization, we opt for the commonly used graph convolution network(GCN)~\cite{kipf2016semi} as our encoders: 
\begin{equation}
\label{eq1}
H^{(l+1)}=\sigma\left(\tilde{D}^{-\frac{1}{2}} \tilde{A} \tilde{D}^{-\frac{1}{2}} H^{(l)} W^{(l)}\right)
\end{equation}
where $\tilde{A}=A+I_{N}$ is the adjacency matrix of the input graph $\mathcal{G}$ with added self-connections, $I_{N}$ is the identity matrix and $\tilde{D}_{i i}=\sum_{j} \tilde{A}_{i j}$ is the degree matrix,
$H^{(l)} \in \mathbb{R}^{N \times D}$ is the representation in $l^{th }$ layer,
$\sigma(\cdot)$ is a non-linear activation function,
$W^{(l)}$ is a trainable weight matrix, which is our final goal  to learn.

\subsection{Graph-structured data augmentations}
Recent successful self-supervised learning approaches in visual domain learn representations by contrasting congruent and incongruent augmentations of images~\cite{grill2020bootstrap}. Nevertheless, unlike images with standard augmentation methods, how to get effective augmentation methods for graph-structured data has no consensus. Considering there are actually two kinds of information in the graph: node information and adjacent information, in this paper, we introduce three kinds of graph data augmentations: node augmentation and graph diffusion network for adjacent matrix augmentation and the combination of them. 
\subsubsection{Node Augmentation}
For the node informtion, considering feature matrix $\mathrm{X} \in \mathbb{R}^{n \times d}$, $n$ is the number of nodes and $d$ is the dimensionality of features. Following~\cite{grand} we use node dropout~(ND) and node feature dropout~(NFD): node dropout denotes randomly zeroes one node's entire features with a pre-defined probability, i.e. dropping the row vectors of $\mathrm{X}$, randomly; while node feature dropout means randomly discards each element of $\mathrm{X}$. The specific formulation of ND and NFD are  
\begin{equation}
\widetilde{\mathrm{X}}_{i}=\frac{\tilde{\epsilon}_{i}}{1-\delta_{nd}} \mathrm{X}_{i}
\end{equation}
\begin{equation}
\widetilde{\mathrm{X}}_{i j}=\frac{\tilde{\epsilon}_{i j}}{1-\delta_{nfd}} \mathrm{X}_{i j}
\end{equation}
where $\delta_{nd}$ and $\delta_{nfd}$ are the dropout probability of ND and NFD  repectively, $\tilde{\epsilon}_{i}$ and  $\tilde{\epsilon}_{i j}$ seperately draws from \emph{Bernoulli}(1 - $\delta_{nd}$), \emph{Bernoulli}(1 - $\delta_{nfd}$). The factor $1/(1-\delta_{nd})$ and $1/(1-\delta_{nfd})$ are to make the perturbed feature matrix $\widetilde{\mathrm{X}}$ equal to $\mathrm{X}$ in expectation. To note that, NFD and ND are only used during training. After training, we use initial node features  to calculate representations.

\subsubsection{Adjacent Matrix Augmentation}
For the adjacent matrix augmentation, we consider graph diffusion networks~\cite{klicpera2019diffusion}. In previous GNNs, there exist two problems: one is that each GNN layer limits the message passing within one-hop neighbors, which is not reasonable in the real world; the other is that edges are often noisy or defined with an arbitrary threshold~\cite{tang2018atomistic}. 
Graph diffusion networks(GDN)~\cite{klicpera2019diffusion} are proposed to tackle the two problems. GDN combines spatial message passing with a sparsified form of graph diffusion which can be regarded as an equivalent polynomial filter. GDC generates a new graph by sparsifying a generalized form of graph diffusion, as a result, GDC can aggregate information from a larger neighborhood rather than only from the first-hop neighbors.

For a graph $\mathcal{G} = \{\mathbf{X}, \mathbf{A}\}$, the generalized graph diffusion is formulated as:
\begin{equation}
\label{eq2}
S=\sum_{k=0}^{\infty} \theta_{k} \mathbf{T}^{k}
\end{equation}
where $\mathbf{T} \in \mathbb{R}^{n \times n}$ denotes the generalized transition matrix, $\theta_{k}$ is the weighting coefficient which determines the ratio of global-local information. In order to guarantee convergence, two conditions are considered: $\sum_{k=0}^{\infty} \theta_{k}=1, \theta_{k} \in[0,1]$; $\lambda_{i} \in[0,1]$ where $\lambda_{i}$ are eigenvalues of $\mathbf{T}$.

Two popular instantiations of the graph diffusion are personalized PageRank~(PPR)~\cite{page1999pagerank} and heat kernel~\cite{kondor2002diffusion}. For an adjacency matrix 
$\mathbf{A} \in \mathbb{R}^{n \times n}$ and a degree matrix $\mathbf{D} \in \mathbb{R}^{n \times n}$, $\mathbf{T}$ in equation~\ref{eq2} is defined as $\mathbf{T}=\mathbf{A} \mathbf{D}^{-1}$, $\theta_{k}=\alpha(1-\alpha)^{k}$ for PPR and 
$\theta_{k}=e^{-t} t^{k} / k !$ for heat kernel, where
$\alpha \in(0,1)$ is teleport probability and $t$ is the diffusion time. 
The specific formulation is showed in equation~\ref{eq_heat}, equation~\ref{eq_ppr}~\cite{hassani2020contrastive}:

\begin{equation}
\label{eq_heat}
\mathbf{S}^{\text {heat }}=\exp \left(t \mathbf{A} \mathbf{D}^{-1}-t\right)
\end{equation}
\begin{equation}
\label{eq_ppr}
\mathbf{S}^{\mathrm{ppr}}=\alpha\left(\mathbf{I}_{n}-(1-\alpha) \mathbf{D}^{-1 / 2} \mathbf{A} \mathbf{D}^{-1 / 2}\right)^{-1}
\end{equation}

In the "augmentation" part of Figure~\ref{fig_whole}, we apply NFD to the first view, and NFD + graph diffusion to the second view. Comparing with the initial graph, we can easily see how these augmentation methods work. For simplicity, we do not scale the augmented node features in Figure~\ref{fig_whole}. We apply different combinations of the augmentation methods above to DGB, and the results and discussions will be in Section 5.

\subsection{Bootstrapping Process}
The bootstrapping process is the core of the DGB model. As is shown in figure \ref{fig_whole}, the online network consists of three steps: one graph convolution network layer called $\mathrm{GNN}_{\theta}$, one multilayer perceptron for projection $\mathrm{MLP}_{\theta}^{\text {pro}}$, and one multilayer perceptron for prediction $\mathrm{MLP}_{\theta}^{\text {pre}}$; similar to the online network, the target network has corresponding $\mathrm{GNN}_{\varepsilon}$ and $\mathrm{MLP}_{\epsilon}^{\text {pro}}$. For simplicity, in the following, we use $\theta$ to denote the parameters of $\mathrm{GNN}_{\theta}$, $\mathrm{MLP}_{\theta}^{\text {pro}}$, and use $\varepsilon$ to denote the parameters of $\mathrm{GNN}_{\varepsilon}$, $\mathrm{MLP}_{\epsilon}^{\text {pro}}$. The reason for using a multilayer perceptron for projection has been proved to improve performances~\cite{chen2020simple}, and we will have a further discussion in the ablation study.

During each epoch training, the parameters of the target network are fixed, and the regression loss between the online network outputs  $q(z)$ and the target network outputs $Z^{\prime}$ are used to update the online network parameters. After one epoch training, the target network parameters $\varepsilon$ are updated using an exponential moving average of the  $\theta$~\cite{lillicrap2015continuous,guo2020bootstrap}. Specifically, after one epoch training, for a given target decay rate $p \in[0,1]$, this paper uses the following updating process:

\begin{equation}
\label{eq_update}
\xi \leftarrow p \xi+(1-p) \theta
\end{equation}
when $p = 1$, the target network will be never updated and remains at a constant value; when $p = 0$, the target network  will be updated to the online network at each step. Therefore, we need a trade-off value for $p$ to update the target network at a proper speed.

\subsection{Training Method}
As is shown in Figure~\ref{fig_whole}, the loss function in DGB is the mean squared error between online network predictions $q\left(z\right)$ and target network projections $z^{\prime}$. Before calculating the error, we first $\ell_{2}$-normalize the predictions and projections:

\begin{equation}
\label{eq_update}
\overline{q}(z) \triangleq q\left(z\right) /\left\|q\left(z\right)\right\|_{2}
\end{equation}

\begin{equation}
\label{eq_update}
\bar{z}^{\prime} \triangleq z^{\prime} /\left\|z^{\prime}\right\|_{2}
\end{equation}
The loss function is:

\begin{equation}
\label{eq_update}
\mathcal{L}_{\theta}^{\mathrm{DGB}} \triangleq\left\|\overline{q}(z)-\bar{z}^{\prime}\right\|_{2}^{2}
\end{equation}
After getting $\mathcal{L}_{\theta}^{\mathrm{DGB}}$, we seperately feed the second view augmentation to the online network and the first view augmentation to the target network to compute $\widetilde{\mathcal{L}}_{\theta}^{\mathrm{DGB}}$. During each epoch training,  $\mathcal{L}_{\theta}^{\mathrm{DGB}}$ + $\widetilde{\mathcal{L}}_{\theta}^{\mathrm{DGB}}$ is used to minimize via stochastic optimization with respect to the parameters of $\mathrm{GNN}_{\theta}$, $\mathrm{MLP}_{\theta}^{\text {pro}}$ and $\mathrm{MLP}_{\theta}^{\text {pre}}$.

After training, we only keep the GNN layer of the online network $\mathrm{GNN}_{\theta}$, and use $\mathrm{GNN}_{\theta}$ to compute the representations for downstream tasks.

\section{Experiments}
\subsection{Datasets}
In this section, we introduce experiments to demonstrate the effectiveness of the DGB model. The datasets in our experiments are three standard benchmark citation network datasets, namely Cora, Citeseet, and Pubmed~\cite{sen2008collective}. In the three datasets, nodes represent documents, edges correspond to citations, and each node has a feature vector corresponding to the bag-of-words representation. Each node can be divided into one of the several classes. More details about the three datasets are in Table~\ref{datasetinformation}. Following the experimental setup in~\cite{kipf2016semi}, we use the same data splits.

\begin{table}[H]
	\caption{Statistics of benchmark datasets}\smallskip
	\centering
	\resizebox{0.7\columnwidth}{!}{
		\smallskip
		\begin{tabular}{ccccc}
			\hline
			Dataset & Nodes & Edges & Features & Classes \\
			\hline
			Cora    & 2708  & 5429  & 1433     & 7       \\
			\hline
			Citeseer& 3327  & 4732  & 3703     & 6       \\ 
			\hline
			Pubmed  & 19717 & 44338 & 500      & 3      \\
			\hline
		\end{tabular}
	}
	\label{datasetinformation}
\end{table}

\subsection{Evaluation Protocol}
The aim of the DGB model is to learn a low dimensional representation of each node based on the node  features and the interactions between nodes. After the training process of the DGB model, we use a linear evaluation to demonstrate the effectiveness of the learned representations. Following DGI~\cite{velickovic2019deep}, we report the mean  classification accuracy on the test nodes after 50 runs of training followed by a linear model. 

To note that, the standard deviation of 50 runs in our experiments is very small so we don't list it in Table~\ref{classification accuracy}.

\subsection{Baselines}
To comprehensively evaluate the proposed DGB model, we compare it with six supervised methods and eight unsupervised methods in Table~\ref{classification accuracy}.

\begin{itemize}
	\item LP~\cite{zhu2003semi} is based on a gaussian random field model.
	\item PLANETOID~\cite{yang2016revisiting} proposes joint training of classification and graph context prediction.
	\item CHEBYSHEV~\cite{defferrard2016convolutional} uses the methods of  graph signal processing.
	\item GCN~\cite{kipf2016semi} generalizes traditional convolution network to graph structure data.
	\item GAT~\cite{velivckovic2017graph} puts attention mechanism into GCN.
	\item GWNN~\cite{xu2019graph} leverages graph wavelet transform instead of graph fourier transform.			
	\item RAW FEATURES method feeds the node features into a logistic regression classifier for training and gives the results on the test features~\cite{peng2020graph}. 
	\item DEEPWALK~\cite{perozzi2014deepwalk} learns representations via random walks on graphs.
	\item EP-B~\cite{duran2017learning} learns label and node representations by exchanging messages between nodes.
	\item GMI~\cite{peng2020graph} maximizes the mutual information between node features and topological structure.
	\item DGI~\cite{velickovic2019deep} maximizes mutual information at graph/patch-level.
	\item{GMNN}~\cite{qu2019gmnn} combines the advantages of the statistical relational learning and  graph neural network to learn node representations.
\end{itemize}
The results of LP and DEEPWALK are taken from~\cite{kipf2016semi}, and the results of RAW FEATURES and CHEBYSHEV are taken from~\cite{peng2020graph} and~\cite{hassani2020contrastive} respectively. The results of other methods are taken from their original papers.

\subsection{Experiments Setup}
All experiments are implemented in Pytorch~\cite{ketkar2017introduction} and conducted on a single Geforce RTX 2080Ti with 11GB memory size.
%For the GCN encoder, we adopt the implementation of GCN layer from the PyTorch-Geometric library~\cite{Fey/Lenssen/2019}. 
We use Glorot initialization~\cite{glorot2010understanding} to initialize the parameters of the model. We perform row normalization on the three datasets for preprocessing. For graph convolution encoder, we use one layer on the three datasets. During training, we use Adam optimizer~\cite{kingma2014adam} with an initial learning rate of 0.001 for the three datasets.
% The target decay rates are 0.82 for Cora, 0.88 for Citeseer and 0.95 for Pubmed.

%In DGB model, the data augmentation methods play an important role in DGB. Consequently, we conduct extensive experiments for different data augmentation combinations in Table~\ref{extra_experiments}. When we finish training the graph convolution network encoder, we find it that feeding diffusion matrix to the encoder can achieve higher performance than adjacent matrix~corresponding to DGB-DIFF in Table~\ref{classification accuracy}. However, to make a fair comparsion with other methods, we also list results with utilizing adjacent matrix, correspongding to DGB-ADJ.

In DGB model, the data augmentation methods play an important role. Consequently, we conduct extensive experiments for different data augmentation combinations in Table~\ref{extra_experiments}. After training, we feed the initial node features with the adjacent matrix to the GCN encoder and get node representations for classification task. 

\begin{table}[H]
	\centering
	\caption{Mean classification accuracy in percent for supervised and unsupervised models on three benchmark datasets. We list the data available to each model during training in the third column. X, A, and Y represent node features, adjacency matrix and node labels, respectively.}\smallskip
	\resizebox{1.\columnwidth}{!}
	{\smallskip
		\begin{tabular}{cllccc}
			\hline
			& METHOD & Available Data & Cora & Citeseer & Pubmed \\
			\hline
			\multirow{6}*{Supervised}
			& LP            & A, Y     & 68.0  & 45.3   & 63.0 \\
			& PLANETOID & X, Y     & 75.7  & 62.9   & 75.7 \\
			& CHEBYSHEV & X, A, Y  & 81.2  & 69.8   & 74.4 \\
			& GCN          & X, A, Y  & 81.5  & 70.3   & 79.0 \\
			& GAT   & X, A, Y  & 83.0$\pm$ 0.7 & 72.5$\pm$0.7 & 79.0$\pm$0.3 \\
			& GWNN           & X, A, Y  & 82.8          & 71.7         & 79.1 \\
			\hline
			\multirow{9}*{Unsupervised}
			& RAW FEATURES  &X   & 56.6$\pm$0.4  & 57.8$\pm$0.2 & 69.1$\pm$0.2 \\
			& DEEPWALK     &X, A& 70.7$\pm$0.6  & 51.4$\pm$0.5 & 74.3$\pm$0.9 \\
			& EP-B         &X, A& 78.1$\pm$1.5  & 71.0$\pm$1.4 & 79.6$\pm$2.1 \\
			& GMI-mean          &X, A& 82.7$\pm$0.2  & 73.0$\pm$0.3 & 80.1$\pm$0.2 \\
			& GMI-adaptive  &X, A& 83.0$\pm$0.3  & 72.4$\pm$0.1 & 79.9$\pm$0.2 \\
			& DGI          &X, A& 82.3$\pm$0.6  & 71.8$\pm$0.7 & 76.8$\pm$0.6 \\
			& GMNN(with $q_{\theta}$) &X, A& 78.1        & 68.0         & 79.3\\
			& GMNN(with $q_{\theta}$ and $p_{\phi}$)  &X, A &82.8 &71.5 & 81.6 \\
			& DGB(ours)                              &X, A&$\mathbf{83.4}$          & $\mathbf{73.9}$         &$\mathbf{81.9}$   \\
			\hline
		\end{tabular}
	}
	\label{classification accuracy}
\end{table} 

\subsection{Experiment Analysis}
From Table~\ref{classification accuracy}, it is apparent that DGB can  perform the best among the recent state-of-the-art methods. For example, on Cora, DGB can improve GMI-adaptive by a margin 0.4\%, and improve GMI-mean by a margin 0.9\% on Citeseer. 

We consider this improvement benefitting from two points: one is that the DGB model can learn from its previous version and does not need well-designed negative examples; the other is that node augmentation methods generate multiple node feature matrixes in different epochs, which alleviates overfitting to some extent and improves the DGB's robustness.

\begin{table}[H]%\normalsize
	\centering
	\caption{Mean node classification accuracy under different combinations of data augmentations. In DGB model, two different augmentations of graph data learn from each other. For the first column, we list one initial graph data and three kinds of graph data augmentations, NO, NODE, ADJ, NODE + ADJ represent no augmentation, node augmentation, adjacent matrix augmentation, and the combination of node and adjacent matrix augmentation. For the second and third columns, we present the specific data augmentation combinations. IN, NFD, ND, ADJ, DIFF denote  initial node, node feature dropout, node dropout, initial adjacent matrix, and diffusion matrix, respectively.}\smallskip
	\resizebox{1.\columnwidth}{!}{
		\begin{tabular}{cllccc} %表格7列 全部居中显示
			\hline
			&First view & Second view & Cora & Citeseer & Pubmed \\
			\hline
			NO &IN + ADJ & IN +ADJ & 59.2 & 51.4 & 62.2 \\
			\hline
			\multirow{5}*{NODE}
			&IN & NFD & 80.6 &71.1 & 77.9\\
			&IN & ND & 76.4 & 63.7 & 77.5 \\
			&NFD & ND & 81.2 & 71.6 & 79.3 \\
			&NFD & NFD& 80.1 & 70.0 & 76.4\\
			&ND  & ND & 76.6 & 64.6 & 74.1 \\
			\hline
			\multirow{1}*{ADJ}
			&DIFF & ADJ & 59.3 & 58.7 & 66.0 \\
			\hline
			\multirow{10}*{NODE + ADJ}
			&IN + ADJ & NFD + DIFF & 80.8 & 71.1 & 79.8 \\
			&IN + ADJ & ND + DIFF & 79.1 & 65.9 & 78.5 \\
			&NFD + ADJ & IN +DIFF & 82.1 & 71.4 & 78.7 \\
			&ND + ADJ & IN + DIFF & 77.9 & 65.4 & 78.4 \\
			&ND + ADJ & ND + DIFF & 79.0 & 65.5 & 77.6 \\
			&NFD + ADJ & NFD + DIFF & 81.4 & 72.3 & 79.0 \\
			&ND + ADJ & NFD + DIFF & 82.3 & 73.3 &  79.4 \\
			&NFD + ADJ &ND + DIFF & $\mathbf{83.4}$ & $\mathbf{73.9}$ & $\mathbf{81.9}$ \\
			\hline
		\end{tabular}
	}
	\label{extra_experiments}
\end{table}

\subsection{Results under different graph data augmentations}
In order to demonstrate the relationship between different graph data augmentations and the performances of the DGB model, we combine different augmentations in DGB and show the results in Table~\ref{extra_experiments}. In node augmentation experiments, we only use the adjacent matrix; while in adjacent augmentation experiments, we use the initial node features without node dropout or node feature dropout. For fairness, the experiments on one dataset in Table~\ref{extra_experiments} are under the same hyperparameters except for the dropout rate. We select the best dropout rate in node dropout and node feature dropout from 0.1 to 0.9 with step 0.1. After training, we feed the feature matrix and adjacent matrix without augmentation to get representations for node classification.

Experimental results in Table~\ref{extra_experiments} show the performances under different graph data augmentations in DGB on Cora, Citeseer, and Pubmed. From Table~\ref{extra_experiments}, we obtain several observations as follows:
\begin{itemize}
	\item The results on the three datasets show that node augmentation and adjacent augmentation can both improve the performances, and combining the two augmentations can achieve higher performances overall.
%	\item For the node augmentation group in the second row, though it gets the worst performances in Table~\ref{extra_experiments}, it does not get trivial solutions. It may be because the slow-moving average mechanism prevents the model falling into trivial solutions.
	\item For node augmentation, it is apparent to see NFD does better in improving performance than ND. We suspect randomly dropping one node's entire features may lose too much information and is difficult to predict for the other view.
	\item There is a clear trend: more combinations of data augmentations bring better performances. For node augmentation group, the combination of NFD \& ND can beat the other combinations on the three datasets. Similarly, for node + adj augmentation group, the combination of NFD + ADJ \& ND + DIFF and ND + ADJ \& NFD + DIFF are better than other combinations. 
	\item Since the graph diffusion brings more connections to the nodes, the combination of ND + DIFF does not hurt the graph information too much compared with ND + ADJ. As a result, the combination of NFD + ADJ \& ND + DIFF is superior to the combination ND + ADJ \& NFD + DIFF.
	\item Though node augmentation and adjacent augmentation can both benefit the model's performance, node augmentation is superior to adjacent augmentation. The results of NODE augmentation group are much better than ADJ augmentation group on all the three datasets. This may be because node augmentations can generate different node feature matrixes in each epoch because of the randomness of the Bernoulli distribution, while the diffusion matrix is the same across different epochs. The randomness improves the performance and robustness of the DGB model.
	%	\item There is a tradeoff between the number of data augmentations and the performances. For instance, for node augmentation, the NFD vs ND group with two node augmentations achieves higher performance than the other group with only one data augmentation. However, the group NFD + ADJ vs NFD +DIFF with two data augmentations is better than the group  ND + ADJ vs NFD + DIFF with three data augmentations.
	%	We do not need to use the most data augmentations, and we need the most suitable data augmentation combination.
\end{itemize}

\subsection{Ablation study}
In this section, we consider the function of a multilayer perceptron for projection in our DGB model and how the bootstrapping mechanism helps the DGB model. We conduct experiments on the three datasets with and without a multilayer perceptron for projection. For analysizing the bootstrapping mechanism, we seperately let $p$=0 and $p$=1. When $p$=0, the target network will copy the parameters of the online network after each epoch training; meanwhile the target network parameters will remain constant values when $p$=1.
\begin{table}[h]
	\caption{Ablation study results}\smallskip
	\centering
	\resizebox{0.7\columnwidth}{!}{
		\smallskip
		\begin{tabular}{cccc}
			\hline
			& Cora & Citeseer & Pubmed\\
			\hline
			DGB & 83.4 & 73.9 & 81.9\\
			\hline
			without projection & 76.7 & 66.9 & 65.6\\
			\hline
			%			without bn & 34.8 & 34.8 & 61.4\\
			%			\hline
			$p$=0 & 82.2    &  72.1    &  80.3   \\
			\hline
			$p$=1 & 74.6    &  62.6    &  69.0   \\
			\hline
		\end{tabular}
	}
	\label{ablation_study}
\end{table}

As is shown in Table~\ref{ablation_study}, the performances of the DGB model with and without MLP have a big difference on the three datasets. Keeping the target network parameters constant or changing them to the online network parameters totally both hurt the performances.

%For Cora, the model with MLP has more than 11\% percent improvement upon the model without MLP; the improvement is about 8\% percent for Pubmed. These differences  demonstrate the effectiveness of the MLP.
\begin{figure}[H]
	\centering
	\includegraphics[width=1.\columnwidth]{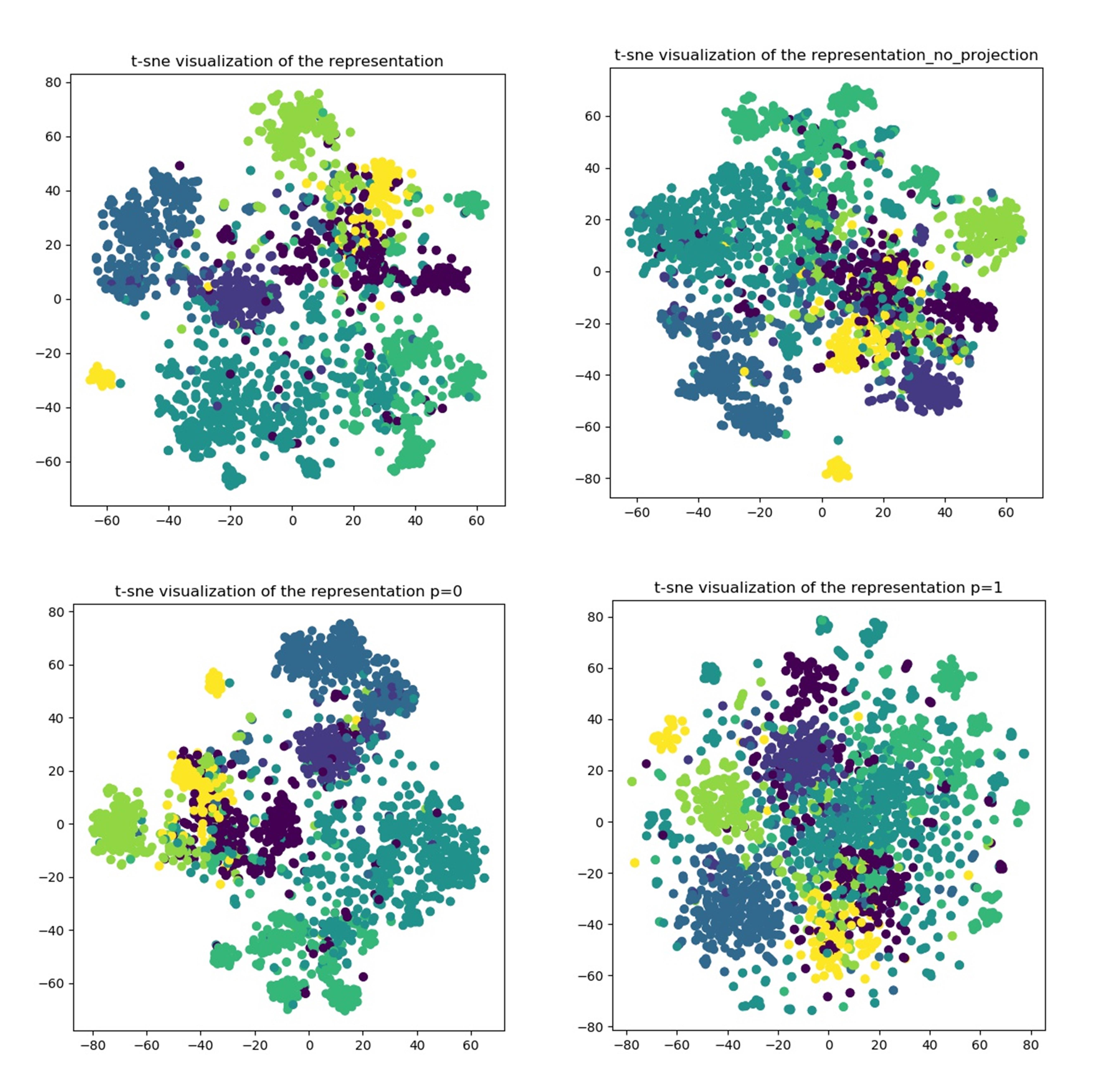} 
	\caption{The visualization of the representations on Cora dataset.}
	\label{visualization}
\end{figure}
We also visualize the node representations of the online network using t-sne~\cite{maaten2008visualizing} on Cora dataset in Figure~\ref{visualization}. In Figure~\ref{visualization}, there are four subgraphs: the representations of the DGB model, the DGB model without a projection layer, the DGB model with $p$=0 and $p$=1.

Comparing the four subgraphs, we can see different classes of the representations without the projection layer become tighter and are not easy to distinguish. When $p$=0, the representations are  better than the representations with $p$=0, but they are all worse than the representations with slow-moving average mechanism. Not changing the target network parameters or changing them totally are both not the best option, and there is a trade-off between the two choices.

\section{Conclusion}
In this paper, we introduce a new self-supervised graph representation learning method DGB. DGB relies on two neural networks: online network and target network, and the input of each neural network is an augmentation of the initial graph. With the help of the bootstrapping process, the online network and target network can learn from each other. As a result, DGB does not need negative examples and can learn in an unsupervised manner. Experiments on three benchmark datasets show DGB is superior to state-of-the-art methods. In addition, we systematically conclude different graph data augmentation methods: node augmentations, adjacent matrix augmentations and the combination of them. We also apply different data augmentation types to DGB and experiment results show how different augmentation methods affect the performances of the DGB. As we only apply DGB to the homogeneous network, we will focus on how to generalize the  DGB model to heterogenous information networks in the future.

\section*{References}
\bibliography{dgb}
\end{document}